\documentclass[11pt,a4paper]{article}
\usepackage[hyperref]{emnlp2018}
\usepackage{times}
\usepackage{latexsym}
\usepackage{graphicx}
\usepackage{booktabs} 
\usepackage{nicefrac}
\usepackage{amsfonts}

\usepackage{url}

\aclfinalcopy 

\author{Yedid Hoshen \\
  Facebook AI Research \\
  \\\And
  Lior Wolf \\
  Facebook AI Research and Tel Aviv University \\}

\date{}

\title{Non-Adversarial Unsupervised Word Translation}

\date{}

\begin{document}
\maketitle
\begin{abstract}
Unsupervised word translation from non-parallel inter-lingual corpora has attracted much research interest. Very recently, neural network methods trained with adversarial loss functions achieved high accuracy on this task. Despite the impressive success of the recent techniques, they suffer from the typical drawbacks of generative adversarial models: sensitivity to hyper-parameters, long training time and lack of interpretability. In this paper, we make the observation that two sufficiently similar distributions can be aligned correctly with iterative matching methods. We present a novel method that first aligns the second moment of the word distributions of the two languages and then iteratively refines the alignment. Extensive experiments on word translation of European and Non-European languages show that our method achieves better performance than recent state-of-the-art deep adversarial approaches and is competitive with the supervised baseline. It is also efficient, easy to parallelize on CPU and interpretable.              
\end{abstract}

\section{Introduction}

Inferring word translations between languages is a long-standing research task. Earliest efforts concentrated on finding parallel corpora in a pair of languages and inferring a dictionary by force alignment of words between the two languages. An early example of this approach is the translation achieved using the Rosetta stone. 

However, if most languages share the same expressive power and are used to describe similar human experiences across cultures, they should share similar statistical properties. Exploiting statistical properties of letters has been successfully employed by substitution crypto-analysis since at least the 9th century. It seems likely that one can learn to map between languages statistically, by considering the word distributions.  As one specific example, it is likely that the set of elements described by the most common words in one language would greatly overlap with those described in a second language. 

Another support for the plausibility of unsupervised word translation came with the realization that when words are represented as vectors that encode co-occurrences, the mapping between two languages is well captured by an affine transformation~\citep{mikolov2013exploiting}. In other words, not only that one can expect the most frequent words to be shared, one can also expect the representations of these words to be similar up to a linear transformation.

A major recent trend in unsupervised learning is the use of Generative Adversarial Networks (GANs) presented by~\citet{goodfellow2014generative}, in which two networks provide mutual training signals to each other: the generator and the discriminator. The discriminator plays an adversarial role to a generative model and is trained to distinguish between two distributions.  Typically, these distributions are labeled as ``real'' and ``fake'', where ``fake'' denotes the generated samples. 

In the context of unsupervised translation~\citep{conneau2017word,zhang2017adversarial,zhang2017earth}, when learning from a source language to a target language, the ``real'' distribution is the distribution of the target language and the ``fake'' one is the mapping of the source distribution using the learned mapping. Such approaches have been shown recently to be very effective when employed on top of modern vector representations of words.

In this work, we ask whether GANs are necessary for achieving the level of success recently demonstrated for unsupervised word translation. Given that the learned mapping is simple and that the concepts described by the two languages are similar, we suggest to directly map every word in one language to the closest word in the other. While one cannot expect that all words would match correctly for a random initialization, some would match and may help refine the affine transformation. Once an improved affine transformation is recovered, the matching process can repeat.

Naturally, such an iterative approach relies on a good initialization. For this purpose we employ two methods. First, an initial mapping is obtained by matching the means and covariances of the two distributions. Second, multiple solutions, which are obtained stochastically, are employed. 

Using multiple stochastic solutions is crucial for languages that are more distant, e.g., more stochastic solutions are required for learning to translate between English and Arabic, in comparison to English and French. Evaluating multiple solutions relies on the ability to automatically identify the true matching without supervision and we present an unsupervised reconstruction-based criterion for determining the best stochastic solution.  

Our presented approach is simple, has very few hyper-parameters, and is trivial to parallelize. It is also easily interpretable, since every step of the method has a clear goal and a clear success metric, which can also be evaluated without the ground truth bilingual lexicon. An extensive set of experiments shows that our much simpler and more efficient method is more effective than the state-of-the-art GAN based method. 

\section{Related Work}

The earlier contributions in the field of word translation without parallel corpora were limited to finding matches between a small set of carefully selected words and translations, and relied on co-occurrence statistics~\citep{Rapp1995} or on similarity in the variability of the context before and after the word~\cite{fung1995compiling}. Finding translations of larger sets of words was made possible in follow-up work by incorporating a seed set of matching words that is either given explicitly or inferred based on words that appear in both languages or are similar in edit distance due to a shared etymology~\citep{Fung1998, Rapp1999, schafer2002inducing, koehn2002learning, haghighi2008learning, irvine2013supervised, xia2016dual,artetxe2017learning}. 

For example,~\citet{koehn2002learning} matched English with German. Multiple heuristics were suggested based on hand crafted rules, including similarity in spelling and word frequency.
A weighted linear combination is employed to combine the heuristics and the matching words are identified in a greedy manner.
~\citet{haghighi2008learning} modeled the problem of matching words across independent corpora as a generative model, in which cross-lingual links are represented by latent variables, and employed an iterative EM method. 

Another example that employs iterations was presented  by~\citet{artetxe2017learning}. Similarly to our method, this method relies on word vector embeddings, in their case the word2vec method~\cite{mikolov2013efficient}. Unlike our method, their method is initialized using seed matches.

Our core method incorporates a circularity term, which is also used in~\cite{xia2016dual} for the task of NMT and later on in multiple contributions in the field of image synthesis~\cite{kim2017learning,CycleGAN2017}. This term is employed when learning bidirectional transformations to encourages samples from either domain to be mapped back to exactly the same sample when translated to the other domain and back. Since our transformations are linear, this is highly related to employing orthogonality as done in~\cite{xing2015normalized,smith2017offline,conneau2017word} for the task of weakly or unsupervised word vector space alignment. \citet{conneau2017word} also employ a circularity term, but unlike our use of it as part of the optimization's energy term, there it is used for validating the solution and selecting hyperparameters.

Very recently,~\citet{zhang2017adversarial,zhang2017earth, conneau2017word} have proposed completely unsupervised solutions. All three solutions are based on GANs. The methods differ in the details of the adversarial training, in the way that model selection is employed to select the best configuration and in the way in which matching is done after the distributions are aligned by the learned transformation.  

Due to the min-max property of GANs, methods which rely on GANs are harder to interpret, since, for example, the discriminator $D$ could focus on a combination of local differences between the distributions. The reliance on a discriminator also means that complex weight dependent metrics are implicitly used, and that these metrics evolve dynamically during training.

Our method does not employ GANs. Alternatives to GANs are also emerging in other domains. For example, generative methods were trained by iteratively fitting random (``noise'') vectors by~\citet{bojanowski2017optimizing}; In the recent image translation work of~\citet{chen2017photographic}, distinguishability between distribution of images was measured using activations of pretrained networks, a practice that is referred to as the ``perceptual loss''~\cite{johnson2016perceptual}. 

\section{Non-Adversarial Word Translation}
\label{sec:method}

We present an approach for unsupervised word translation consisting of multiple parts: (i) Transforming the word vectors into a space in which the two languages are more closely aligned, (ii) Mini-Batch Cycle iterative alignment. There is an optional final stage of batch-based  finetuning.

Let us define two languages $\cal X$ and $\cal Y$, each containing a set of $N_X$ and $N_Y$ words represented by the feature vectors $x_1..x_{N_X}$ and $y_1..y_{N_Y}$ respectively. Our objective is to find the correspondence function $f(n)$ such that for every $x_n$, $f(n)$ yields the index of the $\cal Y$ word that corresponds to the word $x_n$. If a set of possible correspondences is available for a given  word, our objective is to predict one member of this set. In this unsupervised setting, no training examples of $f(n)$ are given.

\subsection{Approximate Alignment with PCA}
\label{subsec:pca}

Each language consists of a set of words each parameterized by a word vector. A popular example of a word embedding method is FastText~\cite{bojanowski2016enriching}, which uses the internal word co-occurrence statistics for each language. These word vectors are typically not expected to be aligned between languages and since the alignment method we employ is iterative, a good initialization is key.

Let us motivate our approach by a method commonly used in 3D point cloud matching. Let $A$ be a set of 3D points and $TA$ be the same set of points with a rotated coordinate system. Assuming non-isotropic distributions of points, transforming each set of points to its principle axes of variations (using PCA) will align the two point clouds.
As noted by~\citet{daras2012investigating}, PCA-based alignment is common in the literature of point cloud matching.

Word distributions are quite different from 3D point clouds: They are much higher dimensional, and it is not obvious a priori that different languages present different ``views'' of the same ``object'' and share exactly the same axes of variation. The success of previous results, e.g. \cite{conneau2017word}, to align word vectors between languages using an orthonormal transformation does give credence to this approach.  
Our method relies on the assumption that many language pairs share some principle axes of variation. The empirical success of PCA initialization in this work supports this assumption.

For each language [$\cal X$, $\cal Y$], we first select the $N$ most frequent word vectors. In our implementation, we use $N=5000$ and employ FastText vectors of dimension $D=300$. We project the word vectors, after centering, to the top $p$ principle components (we use $p=50$). 

\subsection{Mini-Batch Cycle Iterative Closest Point}
\label{subsec:cicp}

Although projecting to the top principle axes of variation would align a rotated non-isotropic point cloud, it does not do so in the general case. This is due to languages having different word distributions and components of variation.

We therefore attempt to find a transformation $T$ that will align every word $x_i$ from language $\cal X$ to a word $y_{f(i)}$ in language $\cal Y$. The objective is therefore to minimize:
\begin{equation}
\label{eq:icp}
\mathop{\mathrm{argmin}}_T \sum_i \min_{f(i)} | y_i - Tx_{f(i)}|
\end{equation}

Eq.~\ref{eq:icp} is difficult to optimize directly and various techniques have been proposed for its optimization. One popular method used in 3D point cloud alignment is Iterative Closest Point (ICP). ICP solves Eq.~\ref{eq:icp} iteratively in two steps.
\begin{enumerate}
\item For each $y_j$, find the nearest $Tx_i$. We denote its index by $f_{y}(j)=i$ \item Optimize for $T$ in $\sum_j \|y_j - Tx_{f_y(j)}\|$
\end{enumerate}

In this work, we use a modified version of ICP which we call Mini-Batch Cycle ICP (MBC-ICP). MBC-ICP learns transformations $T_{xy}$ for $\cal X \to \cal Y$ and $T_{yx}$ for $\cal Y \to \cal X$. We include cycle-constraints ensuring that a word $x$ transformed to the $\cal Y$ domain and back is unchanged (and similarly for every $\cal Y \to \cal X \to \cal Y$ transformation). The strength of the cycle constraints is parameterized by $\lambda$ (we have $\lambda=0.1$). We compute the nearest neighbor matches at the beginning of each epoch, and then optimize transformations $T_{yx}$ and $T_{xy}$ using mini-batch SGD with mini-batch size 128. Mini-batch rather than full-batch optimization greatly increases the success of the method. Experimental comparisons can be seen in the results section. Note we only compute the nearest neighbors at the beginning of each epoch, rather than for each mini-batch due to the computational expense.

Every iteration of the final MBC-ICP algorithm therefore becomes:
\begin{enumerate}
\item For each $y_j$, find the nearest $T_{xy}x_i$. We denote its index by $f_{y}(j)$
\item For each $x_i$, find the nearest $T_{yx}y_j$. We denote its index by $f_{x}(i)$ 
\item Optimize $T_{xy}$ and $T_{yx}$ using mini-batch SGD for a single epoch of $\{x_i\}$ and $\{y_j\}$ on:

$\sum_j \|y_j - T_{xy}x_{f_y(j)}\| + \sum_i \|x_i - T_{yx}y_{f_x(i)}\|$ + $\lambda \sum_i \|x_i - T_{yx}T_{xy}x_i\| + \lambda \sum_j \|y_j - T_{xy}T_{yx}y_j\|$
\end{enumerate}

A good initialization is important for ICP-type methods. We therefore begin with the projected data in which the transformation is assumed to be relatively small and initialize transformations $T_{xy}$ and $T_{yx}$ with the identity matrix. We denote this step PCA-MBC-ICP. 

Once PCA-MBC-ICP has generated the correspondences functions $f_x(i)$ and $f_y(j)$, we run a MBC-ICP on the original 300$D$ word vectors (no PCA). We denote this step: RAW-MBC-ICP. We initialize the optimization using $f_x(i)$ and $f_y(j)$ learned before, and proceed with MBC-ICP. At the end of this stage, we recover transformations $\bar T_{xy}$ and $\bar T_{yx}$ that transform the 300$D$ word vectors from $\cal X \to \cal Y$ and $\cal Y \to \cal X$ respectively.

\textit{Reciprocal pairs:} After several iterations of MBC-ICP, the estimated transformations become quite reliable. We can therefore use this transformation to identify the pairs that are likely to be correct matches. We use the reciprocity heuristic: For every word $y \in \cal Y$ we find the nearest transformed word from the set $\{T_{xy}x | x \in X\}$. We also find the nearest neighbors for the $\cal Y \to \cal X$ transformation. If a pair of words is matched in both $\cal X \to \cal Y$ and $\cal Y \to \cal X$ directions, the pair is denoted reciprocal. During RAW-MBC-ICP, we use only reciprocal pairs, after the 50th epoch (this parameter is not sensitive). 

In summary: we run PCA-MBC-ICP on the $5k$ most frequent words after transformation to principle components. The resulting correspondences $f_x(i)$ and $f_y(j)$ are used to initialize a RAW-MBC-ICP on the original 300D data (rather than PCA), using reciprocal pairs. The output of the method are transformation matrices $T_{xy}$ and $T_{yx}$.

\subsection{Fine-tuning}
\label{subsec:refine}

MBC-ICP is able to achieve very competitive performance without any further finetuning or use of large corpora. GAN-based methods on the other hand require iterative finetuning~\citep{conneau2017word,hoshen2018identifying} to achieve competitive performance. To facilitate comparison with such methods, we also add a variant of our method with identical finetuning to \cite{conneau2017word}. As we show in the results section, fine-tuning European languages typically results in small improvements in accuracy (1-2\%) for our method, in comparison to 10-15\% for the previous work. Following \citep{xing2015normalized, conneau2017word}, fine-tuning is performed by running the Procrustes method iteratively on the full vocabulary of 200k words, initialized with the final transformation matrix from MBC-ICP. The Procrustes method uses SVD to find the optimal orthonormal matrix between $\cal X$ and $\cal Y$ given approximate matches. The new transformation is used to finetune the approximate matches. We run $5$ iterations of successive transformation and matching estimation steps. 

\subsection{Matching Metrics}
\label{subsec:csls}

Although we optimize the nearest neighbor metric, we found that in accordance with~\cite{conneau2017word}, neighborhood retrieval methods such as Inverted Soft-Max (ISF)~\citep{smith2017offline} and Cross-domain Similarity Local Scaling (CSLS) improve final retrieval performance. We therefore evaluate using CSLS. The similarity between a word $x\in \cal X$ and a word $y\in \cal Y$ is computed as $2\cos(T_{xy}x,y)-r(T_{xy}x)-r(y)$, where $r(.)$ is the average cosine similarity between the word and its 10-NN in the other domain.

\begin{figure*}[t]
\centering
\begin{tabular}{ccc}
\includegraphics[width=0.307\linewidth]{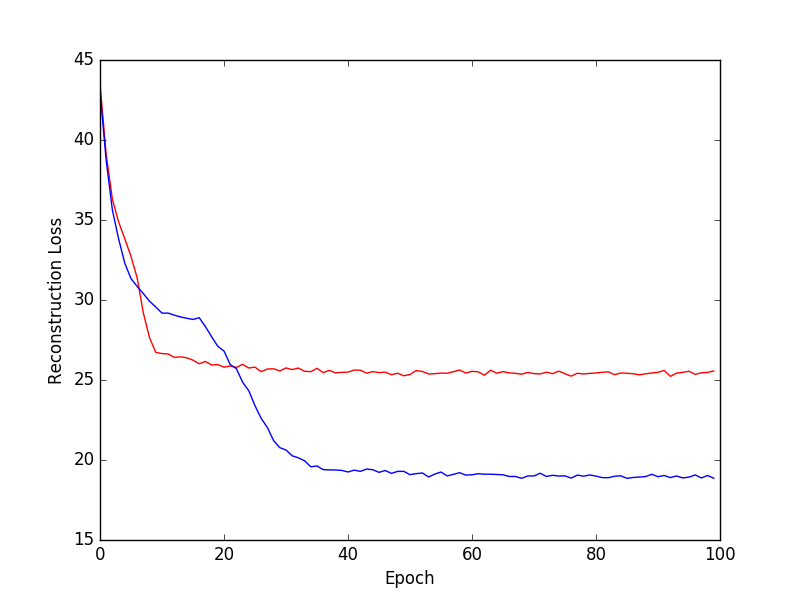} &
\includegraphics[width=0.307\linewidth]{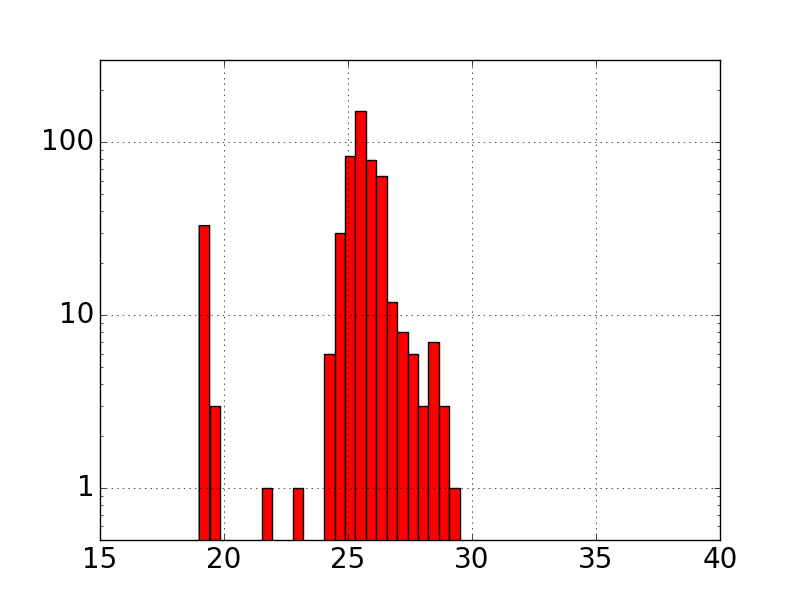} &
\includegraphics[width=0.307\linewidth]{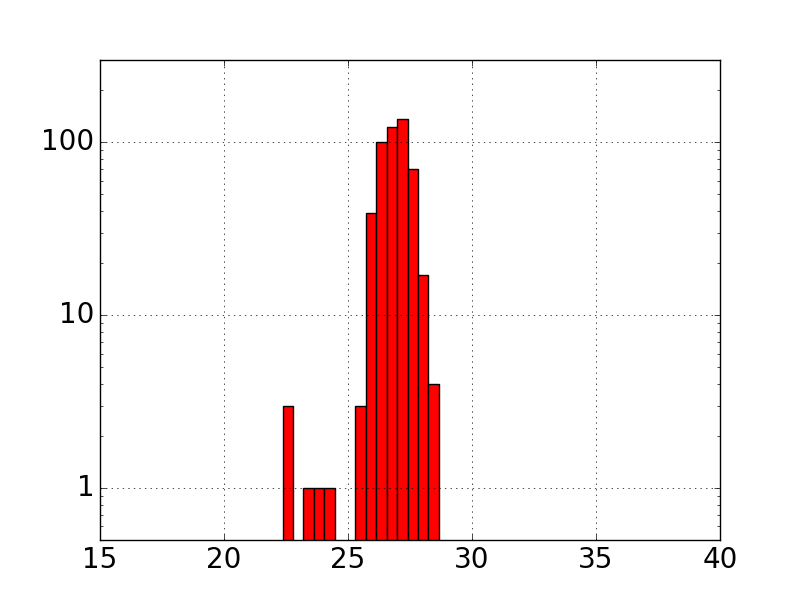} \\
(a) & (b) & (c) \\
\end{tabular}
\caption{(a) Evolution of reconstruction loss as a function of epoch number for successful (Blue) and unsuccessful runs (Red). (b) The final reconstruction loss distribution for En-Fr. (c) A similar histogram for En-Ar.}
\label{fig:converg}
\end{figure*}

\section{Multiple Stochastic Solutions}
\label{subsec:multi_init}

Our approach utilizes multiple stochastic solutions, to provide a good initialization for the MBC-ICP algorithm. There are two sources of stochasticity in our system: (i) The randomized nature of the PCA algorithm (it uses random matrices \cite{liberty2007randomized}) (ii) The order of the training samples (the mini-batches) when training the transformations.

The main issue faced by unsupervised learning in the case of multiple solutions, is either (i) choosing the best solution in case of a fixed parallel run budget, or (ii) finding a good stopping criterion if attempting to minimize the number of runs serially. 

We use the reconstruction cost as an unsupervised metrics for measuring convergence of MBC-ICP. Specifically, we measure how closely every $x \in \cal X$ and $y \in \cal Y$ is reconstructed by a transformed word from the other domain.
\begin{equation}
\sum_j \|y_j - T_{xy}x_{f_y(j)}\| + \sum_i \|x_i - T_{yx}y_{f_x(i)}\|
\end{equation}
Although for isotropic distributions this has many degenerate solutions, empirically we find that values that are significantly lower than the median almost always correspond to a good solution.

The optimization profile of MBC-ICP is predictable and easily lends itself for choosing effective termination criteria. The optimization profile of a successfully converged and non-converging runs are presented in Fig.~\ref{fig:converg}(a).
The reconstruction loss clearly distinguish between the converged and non-converging runs.  Fig.~\ref{fig:converg}(b,c) presents the distribution of final reconstruction costs for 500 different runs for $En$-$Fr$ and $En$-$Ar$.

\section{Experiments}
\label{sec:exp}

We evaluated our method extensively to ensure that it is indeed able to effectively and efficiently  perform unsupervised word translation. As a strong baseline, we used the code and datasets from the MUSE repository by~\citet{conneau2017word}\footnote{https://github.com/facebookresearch/MUSE}. Care was taken in order to make sure that we report these results as fairly as possible: (1) the results from the previous work were copied as is, except for En-It, where our runs indicated better baseline results. (2) For languages not reported, we ran the code with multiple options and report the best results obtained. One crucial option for GAN was whether to center the data or not. From communication with the authors we learned that, in nearly all non-European languages, centering the data is crucial. For European languages, not centering gave better results. For Arabic, centering helps in one direction but is very detrimental in the other. In all such cases, we report the best baseline result per direction. (3) For the supervised baseline, we report both the results from the original paper (in Tab.~\ref{tab:acc_no_ft}) and the results post Procrustes finetuning, which are better (Tab.~\ref{tab:acc}). (4) Esperanto is not available in the MUSE repository at this time. We asked the authors for the data and will update the paper once available. Currently we are able to say (without the supervision data) that our method converges on En-Eo and Eo-En.  

The evaluation concentrated on two aspects of the translation: (i) Word Translation Accuracy measured by the fraction of words translated to a correct meaning in the target language, and (ii) Runtime of the method.

We evaluated our method against the best methods from~\citep{conneau2017word}. The supervised baseline method learns an alignment from 5k supervised matches using the Procrustes method. The mapping is then refined using the Procrustes method and CSLS matching on 200k unsupervised word vectors in the source and target languages. The unsupervised method proposed by~\citep{conneau2017word}, uses generative adversarial domain mapping between the word vectors of the 50k most common words in each language. The mapping is then refined using the same procedure that is used in the supervised baseline.

A comparison of the word translation accuracies before finetuning can be seen in Tab.~\ref{tab:acc_no_ft}. Our method significantly outperforms the method of \citep{conneau2017word} on all of the evaluated European language pairs. Additionally, for these languages, our method performs comparably to the supervised baseline on all pairs except En-Ru for which supervision seems particularly useful. The same trends are apparent for simple nearest neighbors and CSLS although CSLS always performs better. For non-European languages, none of the unsupervised methods succeeds on all languages. We found that the GAN baseline fails on Farsi, Hindu, Bengali, Vietnamese and one direction of Japanese and Indonesian while our method does not succeed on Chinese, Japanese and Vietnamese. We conclude that the methods have complementary strengths, our method doing better on more languages. On languages where both methods succeed, MBC-ICP tends to do much better.   

We present a comparison between the methods after finetuning and using the CSLS metric in Tab.~\ref{tab:acc}. All methods underwent the same finetuning procedure. We can see that our method still outperforms the GAN method and is comparable to the supervised baseline on European languages. Another observation is that on most European language pairs, finetuning only makes a small difference for our method (1-2\%). An unaligned vocabulary of 7.5k is sufficient to achieve most of the accuracy. This is in contrast with the GAN, that benefits greatly from finetuning on 200k words. Non-European language and English pairs are typically more challenging, finetuning helps much more for all unsupervised methods.

It is interesting to examine the languages on which each method could not converge. They typically fall into geographical and semantic clusters. The GAN method failed on Arabic and Hebrew, Hindu, Farsi and Bengali. Whereas our method failed on Japanese and Chinese. We suspect that different statistical properties favor each method.

\begin{table}[t]
\begin{small}
  \centering
  \caption{Comparison of word translation accuracy (\%) - without finetuning. Bold: best unsupervised method.}

    \begin{tabular}{lcccccc}
    \toprule
     \multicolumn{1}{l}{Pair} &\multicolumn{2}{l}{Supervised} & \multicolumn{4}{c}{Unsupervised} \\
      \cmidrule(l){4-7}
	 &  \multicolumn{2}{c}{Baseline} & \multicolumn{2}{c}{GAN} & \multicolumn{2}{c}{Ours}     \\  
    \cmidrule(l){2-3}  \cmidrule(l){4-5} \cmidrule(l){6-7}
	 & nn & csls & nn & csls & nn & csls  \\  
    \midrule
    \multicolumn{7}{c}{European Languages} \\
    \midrule
    En-Es & 77.4& 81.4& 69.8& 75.7&  75.9 & \textbf{81.1}\\
    Es-En & 77.3& 82.9& 71.3& 79.7&  76.0 & \textbf{82.1}\\
    \midrule
    En-Fr & 74.9& 81.1& 70.4& 77.8&  74.8 & \textbf{81.5} \\
    Fr-En & 76.1& 82.4& 61.9& 71.2&  75.0 & \textbf{81.3} \\
    \midrule
    En-De & 68.4& 73.5& 63.1& 70.1&  66.9 & \textbf{73.7} \\
    De-En& 67.7& 72.4 & 59.6& 66.4 & 67.1 & \textbf{72.7} \\
    \midrule
    En-Ru & 47.0& 51.7 & 29.1& 37.2&  36.8 & \textbf{44.4} \\
    Ru-En & 58.2& 63.7 & 41.5& 48.1 &  48.4 & \textbf{55.6} \\
    \midrule
    En-It & 75.7& 77.3 & 54.3 & 65.1 &  71.1 & \textbf{77.0} \\
    It-En & 73.9 & 76.9 & 55.0 & 64.0 &  70.4 & \textbf{76.6} \\
    \midrule
    \multicolumn{7}{c}{Non-European Languages} \\
    \midrule
    En-Fa & 25.7 & 33.1 & * & * &  19.6 & \textbf{29.0} \\
    Fa-En & 33.5 & 38.6 &  * & * &  28.3 & \textbf{28.3} \\
    \midrule
    En-Hi & 23.8 & 33.3 & * & * &  19.4 & \textbf{30.3} \\
    Hi-En & 34.6 & 42.8 &  * & * &  30.5 & \textbf{38.9} \\
    \midrule
    En-Bn & 10.3 & 15.8 & * & * & 9.7  & \textbf{13.5} \\
    Bn-En & 21.5 & 24.6 &  * & * &  7.1 & \textbf{14.5} \\
    \midrule
    En-Ar & 31.3 & 36.5 & 18.9 & 23.5 &  26.9 & \textbf{33.3} \\
    Ar-En & 45.0 & 49.5 & 28.6 & 31 & 39.8 & \textbf{45.5} \\
    \midrule
    En-He & 10.3 & 15.8 & 17.9 & 22.7 &  31.3 & \textbf{38.9} \\
    He-En & 21.5 & 24.6 & 37.3  & 39.1 &  43.4 & \textbf{50.8} \\
    \midrule
    En-Zh & 40.6 & 42.7 & 12.7 & \textbf{16.0} &  * & * \\
    Zh-En & 30.2 & 36.7 & 18.7  & \textbf{25.1} &  * & * \\
    \midrule
    En-Ja & 2.4 & 1.7 & * & * &  * & * \\
    Ja-En & 0.0 & 0.0 & 3.1  & \textbf{3.6} &  * & * \\
    \midrule
    En-Vi & 25.0 & 41.3 & * & * &  * & * \\
    Vi-En & 40.6 & 55.3 &  * & * &  * & * \\
    \midrule
    En-Id & 55.3 & 65.6 & 18.9 & 23.5 &  39.4 & \textbf{57.1} \\
    Id-En & 58.3 & 65.0 &  * & * &  37.1 & \textbf{58.1} \\
	 \bottomrule 
    \end{tabular}
\raggedright{*Failed to converge}
\label{tab:acc_no_ft}
\end{small}
\end{table}

\begin{table}[t]
\begin{small}
  \centering
  \caption{Word translation accuracy (\%) - after finetuning and using CSLS. Bold: best unsupervised methods.}
\begin{tabular}{lccc}
    \toprule
    \multicolumn{1}{l}{Pair}& \multicolumn{1}{l}{Supervised} & \multicolumn{2}{c}{Unsupervised} \\
     \cmidrule(l){3-4} 
    & \multicolumn{1}{c}{Baseline} & GAN &   Ours     \\    
    \midrule
    \multicolumn{4}{c}{European Languages} \\
    \midrule
    En-Es & 82.4 & 81.7&  \textbf{82.1}\\
    Es-En & 83.9 & 83.3&  \textbf{84.1}\\
    \midrule
    En-Fr & 82.3 & \textbf{82.3} &  \textbf{82.3} \\
    Fr-En & 83.2 & 82.1&  \textbf{82.9} \\
    \midrule
    En-De & 75.3 & 74.0&  \textbf{74.7} \\
    De-En & 72.7 & 72.2&  \textbf{73.0} \\
    \midrule
    En-Ru & 50.7 & 44.0&  \textbf{47.5} \\
    Ru-En & 63.5 & 59.1&  \textbf{61.8} \\
    \midrule
    En-It & 78.1 & 76.9&  \textbf{77.9} \\
    It-En & 78.1 & 76.7&  \textbf{77.5} \\
    \midrule
    \multicolumn{4}{c}{Non-European Languages} \\
    \midrule
    En-Fa & 32.6 & * &  \textbf{34.6} \\
    Fa-En & 40.2 & * &  \textbf{41.5} \\
    \midrule
    En-Hi & 34.5 & *&  \textbf{34.6} \\
    Hi-En & 44.8 & * &  \textbf{44.5} \\
    \midrule
    En-Bn & 16.6 & * &  \textbf{14.7} \\
    Bn-En & 24.1 & * &  \textbf{21.9} \\
    \midrule
    En-Ar & 34.5  & \textbf{35.3}&  35.1 \\
    Ar-En & 49.7 & 49.7 &  \textbf{50.6} \\
    \midrule
    En-He & 41.1 & \textbf{41.6}& 40.5  \\
    He-En & 54.9 & 52.6 &  \textbf{52.9} \\
    \midrule
    En-Zh & 42.7  & \textbf{32.5} &  * \\
    Zh-En & 36.7 & \textbf{31.4} &  * \\
    \midrule
    En-Ja & 1.7 & * &  * \\
    Ja-En & 0.0 & \textbf{4.2} &  * \\
    \midrule
    En-Vi & 44.6 & * &  * \\
    Vi-En & 56.9 & * &  * \\
    \midrule
    En-Id & 68.0 & 67.8 & \textbf{68.0} \\
    Id-En & 68.0 & 66.6 &  \textbf{68.0} \\
	 \bottomrule 
    \end{tabular} \\
\raggedright{*Failed to converge}
\label{tab:acc}
\end{small}
\end{table}

\begin{table}[t]
  \begin{small}
  \caption{En-Es accuracy with and without PCA}
  \label{tab:full_exact}
\begin{center}
\begin{tabular}{lcc}
    \toprule
	Method & En-Es & Es-En      \\    
    \midrule
    No PCA & 0.0\% & 0.0\% \\
    With 300 PCs & 0.0\%& 0.0\%\\
    With 50 PCs & \textbf{82.2\%} & \textbf{83.8\%} \\
	 \bottomrule 
    \end{tabular}
    \end{center}
\label{tab:pca}
\end{small}
\end{table}

\begin{table}[t]
\begin{small}

  \caption{Fraction of converging runs per stochasticity.}
\begin{center}\begin{tabular}{lcc}
    \toprule
	Method & En-Es & En-Ar      \\    
    \midrule
    No randomization & 0.0\% & 0.0\% \\
    Randomized Ordering & 0.0\% & 0.0\% \\
    Randomized PCA & 9.8\% & 0.0\% \\
    Randomized Ordering + PCA & \textbf{16.8}\% & \textbf{1.2\%} \\
	 \bottomrule 
    \end{tabular}
    \end{center}
\label{tab:random}
\end{small}
\end{table}

We also compare the different methods in terms of training time required by the method. We emphasize that our method is trivially parallelizable, simply by splitting the random initializations between workers. The run time of each solution of MBC-ICP is 47 seconds on CPU. The run time of all solutions can therefore be as low as a single run, at linear increase in compute resources. As it runs on CPU, parallelization is not expensive. The average number of runs required for convergence depends on the language pair (see below, Fig.~\ref{fig:stats}). We note that our method learns translations for both languages at the same time.

The current state-of-the-art baseline by \cite{conneau2017word} requires around 3000 seconds on the GPU. It is not obvious how to parallelize such a method efficiently. It requires about 30 times longer to train than our method (with parallelization) and is not practical on a multi-CPU platform. The optional refinement step requires about 10 minutes. The performance increase of refinement for our method are typically quite modest and can be be skipped at the cost of 1-2\% in accuracy, the GAN however requires finetuning to obtain competitive results. Another obvious advantage is that our method does not require a GPU.

\textit{Implementation:} We used 100 iterations for the PCA-MBC-ICP stage on 50 PCA word vectors. This was run in parallel over 500 stochastic solutions. We selected the solution with the smallest unsupervised reconstitution criterion. This solution was used to initialize RAW-MBC-PCA, which we run for 100 iterations on the raw word vectors. The latter 50 iterations of RAW-MBC-ICP were carried out with only reciprocal pairs contributing to the optimization. Results were typically not sensitive to hyper-parameter choice, although longer optimization generally resulted in better performance.

\paragraph{Ablation Analyses} 
There are three important steps for the convergence of the ICP step: (i) PCA, (ii) Dimensionality reduction, (iii) Multiple stochastic solutions. In Tab.~\ref{tab:pca} we present the ablation results on the En-Es pair with PCA and no dimensionality reduction, with only the top 50 PCs and without PCA at all (best run out of 500 chosen using the unsupervised reconstruction loss). We can observe that the convergence rate without PCA and with PCA but without dimensionality reduction is much lower than with PCA, the best run without PCA has not succeeded in obtaining a good translation. This provides evidence that both PCA and dimensionality reduction are essential for the success of the method.

\begin{figure*}[t]
\centering
\begin{tabular}{ccc}
\includegraphics[width=0.32\linewidth]{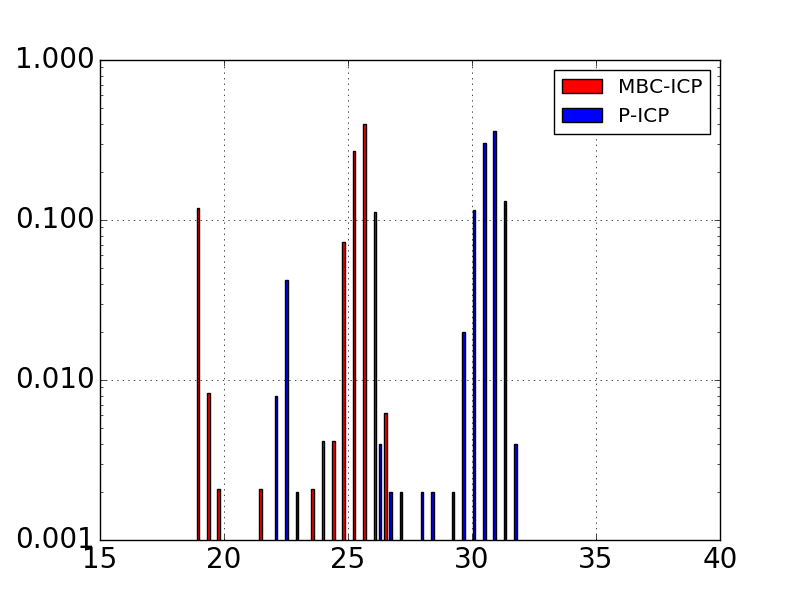}&
\includegraphics[width=0.32\linewidth]{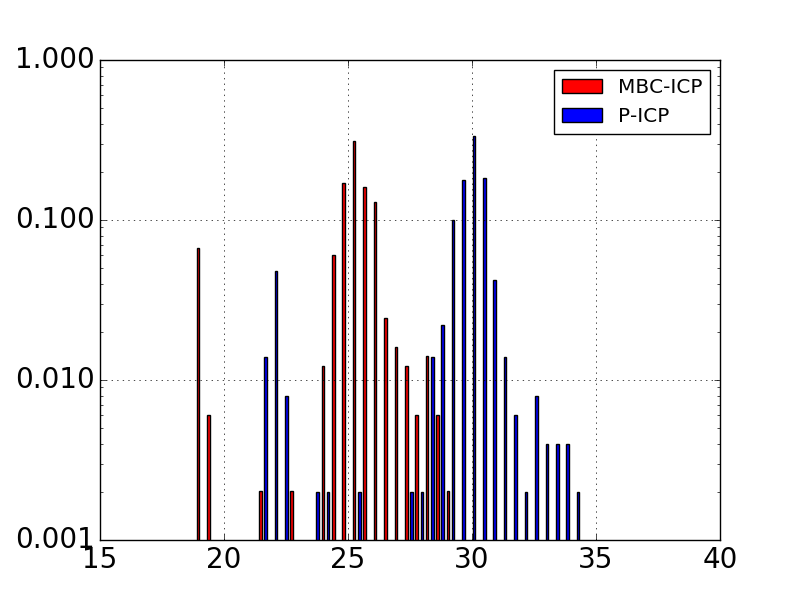}&
\includegraphics[width=0.32\linewidth]{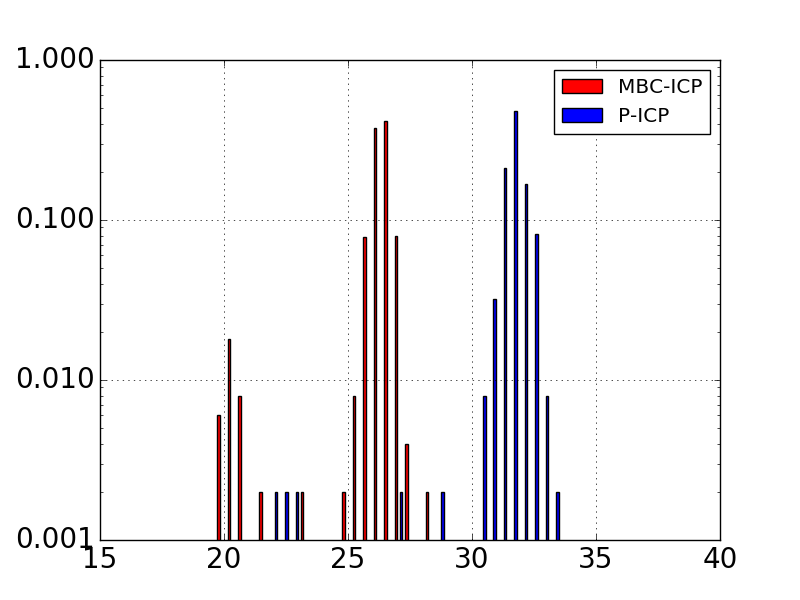}\\
Es & Fr & De \\
\includegraphics[width=0.32\linewidth]{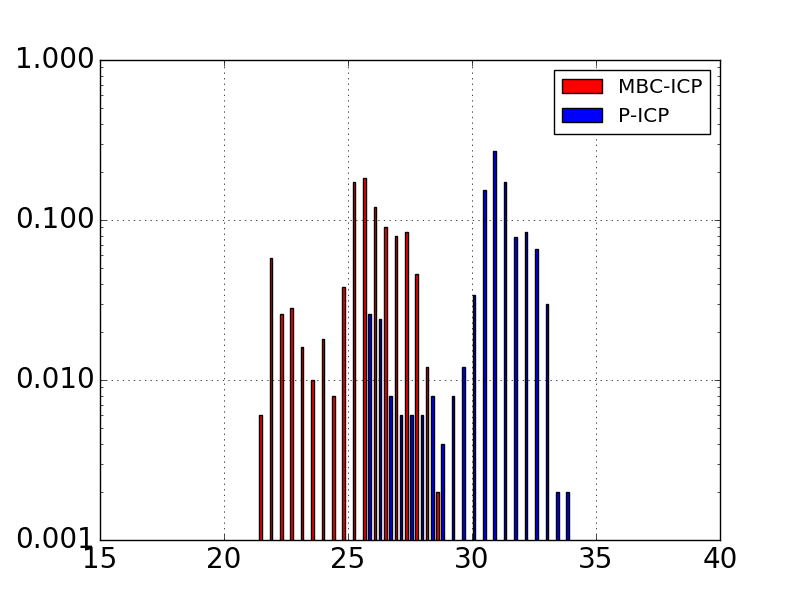}&
\includegraphics[width=0.32\linewidth]{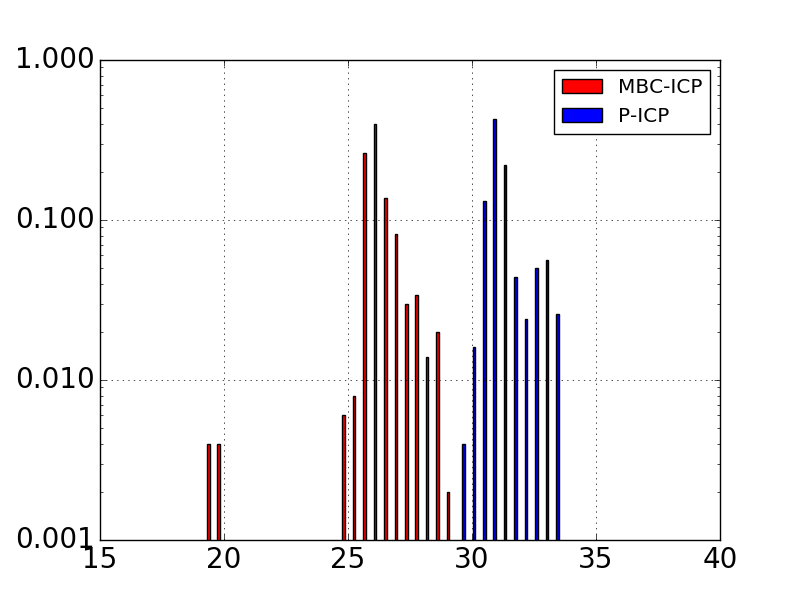}&
\includegraphics[width=0.32\linewidth]{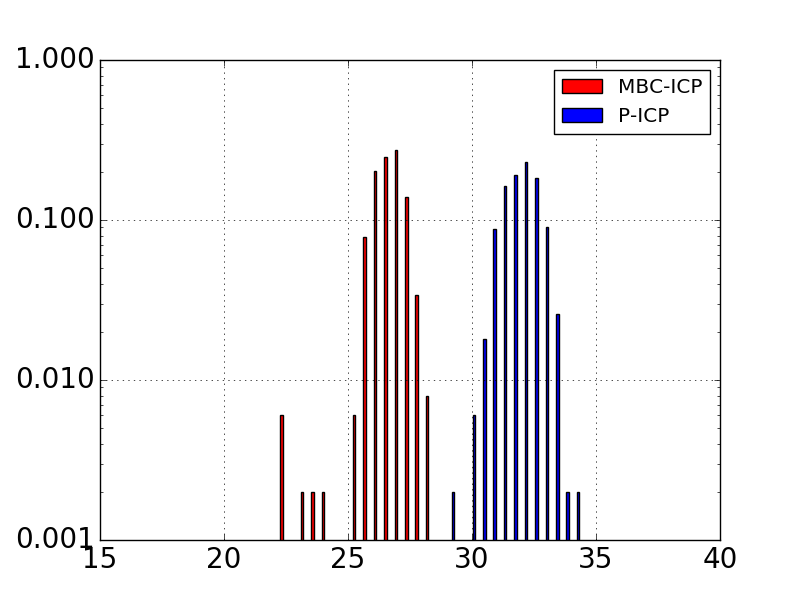}\\
Ru & It & Ar \\
\end{tabular}
\caption{Histograms of the Reconstruction metric across 500 ICP runs for MBC-ICP (Red) and P-ICP (Blue). The comparison is shown for En-Es, En-Fr, En-De, En-Ru, En-It, En-Ar. On average MBC-ICP converges to much better minima. We can observe that MBC-ICP has many more converging runs than P-ICP. In fact for En-It and En-Ar, P-ICP does not converge even once in 500 runs.}
\label{fig:stats}
\end{figure*}
We experimented with the different factors of randomness between runs, to understand the causes of diversity allowing for convergence in the more challenging language pairs (such as En-Ar). We performed the following four experiments: i. Fixing PCA and Batch Ordering. ii. Fixing all data batches to have the same ordering in all runs, iii. Fix the PCA bases of all runs, iv. Let both PCA and batch ordering vary between runs. 

Tab.~\ref{tab:random} compares the results on En-Es and En-Ar for the experiments described above. It can be seen that using both sources of stochasticity is usually better. Although there is some probability the PCA will result in aligned principle components between the two languages, this usually does not happen and therefore using stochastic PCA is highly beneficial.

\paragraph{Convergence Statistics} In Fig.~\ref{fig:stats} we present the statistics for all language pairs with Procrustes-ICP (P-ICP) vs MBC-ICP. In P-ICP, we first calculate the matches for the vocabulary, and then perform a batch estimate of the transformation using the P-ICP method (starting from PCA word vectors and $T_{xy}$ initialized at identity). The only source of stochasticity in P-ICP is the PCA where in MBC-ICP the order of mini-batches provides further stochasticity. Adding random noise to the mapping initialization was not found to help. Each plot shows the histogram in log space for the number of runs that achieved unsupervised reconstruction loss within the range of the bin. The converged runs with lower reconstruction values typically form a peak which is quite distinct from the non-converged runs allowing for easy detection of converged runs. The rate of convergence generally correlates with our intuition for distance between languages (En-Ar much lower than En-Fr), although there are exceptions.

MBC-ICP converges much better than P-ICP: For the language pairs with a wide convergence range (En-Es, En-Fr, En-Ru) we can see that MBC-ICP converged on many more runs than P-ICP. For the languages with a narrow convergence range (En-Ar, En-It), P-ICP was not able to converge at all. We therefore conclude that the mini-batch update and batch-ordering stochasticity increase the convergence range and is important for effective unsupervised matching.

\section{Conclusions}
\label{sec:conc}

We have presented an effective technique for unsupervised word-to-word translation. Our method is simple and non-adversarial. We showed empirically that our method outperforms current state-of-the-art GAN methods in terms of pre and post finetuning word translation accuracy. Our method runs on CPU and is much faster than current methods when using parallelization. This will enable researchers from labs that do not possess graphical computing resources to participate in this exciting field. The proposed method is interpretable, i.e. every stage has an intuitive loss function with an easy to understand objective.

It is interesting to consider the relative performance between language pairs. Typically more related languages yielded better performance than more distant languages (but note that Indonesian performed better than Russian when translated to English). Even more interesting is contrasting the better performance of our method on West and South Asian languages, and GAN's better performance on Chinese.

Overall, our work highlights the potential benefits of considering alternatives to adversarial methods in unsupervised learning. 

\clearpage
\bibliography{icp}
\bibliographystyle{acl_natbib}

\end{document}